# Deep learning-based hyperspectral image reconstruction for quality assessment of agro-product

Md. Toukir Ahmed, Ocean Monjur, Mohammed Kamruzzaman∗

Department of Agricultural and Biological Engineering, University of Illinois at Urbana-Champaign, Urbana, IL 61801, USA.

Abstract

Hyperspectral imaging (HSI) has recently emerged as a promising tool for many agricultural applications; however, the technology cannot be directly used in a real-time system due to the extensive time needed to process large volumes of data. Consequently, the development of a simple, compact, and cost-effective imaging system is not possible with the current HSI systems. Therefore, the overall goal of this study was to reconstruct hyperspectral images from RGB images through deep learning for agricultural applications. Specifically, this study used Hyperspectral Convolutional Neural Network - Dense (HSCNN-D) to reconstruct hyperspectral images from RGB images for predicting soluble solid content (SSC) in sweet potatoes. The algorithm accurately reconstructed the hyperspectral images from RGB images, with the resulting spectra closely matching the ground-truth. The partial least squares regression (PLSR) model based on reconstructed spectra outperformed the model using the full spectral range, demonstrating its potential for SSC prediction in sweet potatoes. These findings highlight the potential of deep learning-based hyperspectral image reconstruction as a low-cost, efficient tool for various agricultural uses.

∗ *Corresponding author. E-mail:* mkamruz1@illinois.edu, *Tel:* + 1 217-265-0423, *Website:* https://abe.illinois.edu/directory/mkamruz1

***Keywords:*** *Hyperspectral Imaging, genetic algorithm, PLSR, image reconstruction, deep learning*

1. Introduction

Hyperspectral imaging (HSI) is a useful technique for comprehensive and non-invasive product quality assessment. Unlike conventional imaging techniques, HSI employs an accumulation of the spectral signature of each spatial location with hundreds of narrow bands (Fu et al., 2021; Kamruzzaman et al., 2016). By combining these extensive spatial and spectral data, HSI creates hypercubes. These hypercubes provide relevant qualitative and quantitative information that may be used to determine the chemical composition, material qualities, and other physical characteristics of each object in an image (Blas Saavedra et al., 2024; Rodríguez-Ortega et al., 2023). Consequently, HSI holds significant promise across diverse applications, including in agriculture, medical imaging, and geology (Ma et al., 2019; Wang et al., 2021; Zhao et al., 2020). In the realm of agriculture, hyperspectral images carry immense potential, particularly concerning the assessment of agro-product quality (Erdogdu, 2023; Pullanagari and Li, 2021; Ryu et al., 2024). HSI is a powerful tool in precision agriculture, helping in data-driven decisions by identifying chlorophyll and nitrogen levels, tracking plant health, spotting diseases, categorizing plant seeds, and sustainably optimizing yield management (Sethy et al., 2022). Despite enormous advantages, high cost, vast and complex data, low spatial resolution, environmental sensitivity, less portability, and low imaging speed make HSI inaccessible for some applications (Fang et al., 2018; Li et al., 2017; Signoroni et al., 2019).

Deep learning has revolutionized data analysis and led to remarkable breakthroughs in image analysis, such as object classification, detection, segmentation, and image generation (K.S. et al., 2023; Mahanti et al., 2022; Minaee et al., 2021). One interesting area of deep learning is hyperspectral image reconstruction from red, green, and blue (RGB) images (Xiong et al., 2017; Zhang et al., 2022). Such reconstruction enables obtaining rich spectral and spatial information of hyperspectral images while using easily accessible RGB images. Reconstructing hyperspectral data from RGB images poses a significant challenge due to the loss of crucial details during integration. Such challenges arise because RGB images capture only a limited portion of the electromagnetic spectrum, resulting in substantial information loss (Shi et al., 2018). However, deep learning algorithms have the potential to overcome existing challenges and to perform desirable reconstruction even with noisy RGB images (Zhang et al., 2022). A desirable hyperspectral image reconstruction from RGB images may significantly broaden the accessibility and applicability of hyperspectral data, enabling more precise analysis and decision-making across a wide range of fields. Such an innovative approach may unlock new possibilities for HSI for research and the potential to revolutionize quality assessment in the food and agriculture industry.

Sweet potatoes (*Ipomoea batatas* L.) offer a wide range of industrial applications, from food processing and textiles to biodegradable plastics and biofuels (Ahmed et al., 2024; He et al., 2023). Sweet potatoes are used as a base ingredient in producing a wide range of processed foods, including sweet potato fries, chips, purees, and snacks. The vast usage of sweet potatoes in the food industry and their nutritional superiority play a vital role in making a potential replacement for raw food materials. However, the taste, nutritional value, marketability, and processing suitability of sweet potatoes significantly depend on their soluble solid content (SSC) (Heo et al., 2021). The conventional experimental processes for SSC analysis are often time-consuming, invasive, and labor-intensive (Sanchez et al., 2020; Shao et al., 2020). Therefore, the integration of deep learning-based hyperspectral image reconstruction from the widely available RGB images would be an innovative approach for rapid, non-destructive prediction of SSC.

Notably, limited works in the literature involve HSI reconstruction for agro-product quality determination (Yang et al., 2024; Zhao et al., 2020). Zhao et al. (2020) applied HSCNN-R reconstruction algorithm to assess the SSC of tomatoes using only 36 samples, achieving an $R^2$ of 0.51. Recently, Yang et al. (2024) used the pre-trained MST++ algorithm to develop inversion models for the psychological parameters of rice, with an $R^2$ of 0.40 for predicting the SPAD value. The findings of the existing studies pave the way for the application of reconstruction algorithms with a deeper network for precise reconstruction of agro-product quality assessment. Therefore, this study aims to effectively train the HSCNN-D reconstruction algorithm for accurately reconstructing hyperspectral images and spectra. Furthermore, the performance of the PLSR model, both on the full spectral range and the selected wavelengths, was compared with the PLSR model applied to the reconstructed spectra. The outcomes of this study may have profound implications for the quality assessment of agricultural products, harvesting decisions and opening new possibilities for HSI in real-world scenarios.

2. Materials and methods

2.1 Sample preparation

This study involved experimentation with the storage roots of three sweet potato cultivars: "Bayou Belle," "Murasaki," and "Orleans." These samples were collected from the Louisiana State University AgCenter sweet potato research station. A total of 141 sweet potatoes, devoid of any defects, were selected across the three cultivars (Bayou Belle - 50, Murasaki - 40, and Orleans - 51) for this investigation. After their collection, all samples were stored at a room temperature of 25°C for a duration of 24 hours before the commencement of image acquisition. Later, the samples were subjected to chemical reference analysis.

2.2 Reference data collection

After the hyperspectral image acquisition, the SSC of each sweet potato sample was determined. This measurement was executed using an Atago Digital Refractometer (Brix 0–32%, Atago, Tokyo, Japan) (Li et al., 2013; Shao et al., 2020). Initially, the flesh of a sample was separated from its outer shell, followed by manual extraction of juice through squeezing. Subsequently, the obtained juice was carefully positioned on the prism plate of the refractometer, and the corresponding reading was recorded. It is important to note that consistent and uniform lighting conditions were maintained throughout this measurement process. The determination of SSC for a particular sample was ultimately derived from the averaging of three replicates of measurements conducted per sample.

2.3 Image acquisition and correction

Preceding the outlined reference analysis, hyperspectral images of sweet potato samples were captured utilizing a Specim V10E IQ line-scan hyperspectral camera designed for visible near-infrared (VNIR) exploration (Specim, Spectral Imaging Ltd., Oulu, Finland). The camera assembly encompassed a lens, imaging spectrograph, and CMOS sensor. Operated with Specim IQ Studio software, the handheld Specim IQ camera featured a 4.3-inch touch screen and 13 buttons for control. Each sample was imaged twice at different viewing angles. The camera boasted 512 spatial pixels and 204 spectral bands, with a 7 nm spectral resolution spanning 400-1000 nm. This consistent configuration generated 2-D square images at a resolution of 512×512 pixels, contributing to developing a 3-D hypercube with dimensions 512×512×204, harboring an extensive dataset of over 53 million data points. Symmetrically positioned 750 W tungsten halogen lamps (ARRILITE 750 Plus, ARRI, Germany) ensured even field-of-view illumination. Addressing non-uniform illumination, hyperspectral imagery of a 99% reflective white image was obtained. For sensor noise mitigation, reference images of a dark background (~0% reflectance) were acquired by blocking light when the lamps were off. These references, comprising the white and dark current images, facilitated noise reduction in the initial unprocessed hyperspectral images by applying Eq. 1.

$$I_R = \frac{I_{R0} - I_D}{I_W - I_D} \quad (1)$$

Where $I_W$ represents the white reference image, $I_D$ represents the dark current image, $I_{R0}$ represents the unprocessed hyperspectral images, and $I_R$ is the corrected image. Band-interleaved-by-line (BIL) format was used to save the acquired hyperspectral images.

2.4 Image segmentation and extraction of spectral data

Image segmentation is one of the essential preliminary stages to derive spectral information from hyperspectral images (Ahmed et al., 2024). The primary objective of image segmentation is to identify regions of interest (ROI) of the tested sample by constructing a mask to distinguish them from the background. Using the differences between two reflectance bands (602 nm and 452 nm), a mask was created by segmenting the sweet potatoes from the image background following image acquisition. The mask was subsequently used as the region of interest (ROI) for extracting reflectance spectral information from the hyperspectral images. The reflectance spectra from the ROIs were then computed by averaging all pixel spectra within the ROI to produce a single mean spectrum for each hyperspectral image. Finally, the spectra for a sample were computed by averaging two spectra from the corresponding images of the same sample.

2.5 Features selection from the extracted spectra

Spectral signatures may contain unnecessary and redundant information throughout the entire spectral region, hurting calibration precision. By judiciously selecting a few informative features from many closely spaced spectral features, the prediction capacity of multivariable analysis can be improved, and the model's complexity can be reduced (Ahmed and Kamruzzaman, 2024; Kamruzzaman et al., 2022). Therefore, considerable efforts have been made to devise and evaluate effective feature selection methods. However, because each experimental dataset is unique and intricate, no singular optimal method applies to all datasets (Pu et al., 2015). Following the extraction of spectral data, the evolution-based feature selection method, namely the genetic algorithm (GA) (Costa et al., 2020) was employed to select some significant wavelengths for the SSC prediction of each sample. This collection of prevalent wavelengths was subsequently deployed to construct the data hypercube for the algorithm for hyperspectral image reconstruction.

2.6 Hyperspectral image reconstruction

2.6.1 Data hypercube creation

This step used the selected feature set to create the data hypercube. Firstly, the mask created in section 2.4 was imposed again on the hyperspectral image. Next, instead of choosing all 204 wavelengths, a few wavelengths from the image were selected. Thus, the dimension of each hypercube was $512 \times 512 \times \lambda$ ($\lambda$ denotes the selected wavelengths). As each sample was imaged twice at different viewing angles, for 141 samples, 282 such hypercubes were constructed and later fed to the hyperspectral image reconstruction algorithm as label information.

2.6.2 Input RGB image preparation

Specim IQ VNIR camera rendered a set of RGB images from hyperspectral images according to CIE Standard Illuminant D65 with the CIE 1931 2° Standard Observer using gamma correction ($\gamma = 1.4$). The generated RGB images were in the PNG (Portable Network Graphics) format. The HSCNN-D reconstruction algorithm, however, only accepts RGB input in JPEG (Joint Photographic Experts Group) format. Therefore, the PNG files were converted to JPEG format before the RGB images were input into the reconstruction algorithm. The same binary mask obtained in section 2.4 was applied to them.

2.6.3 The HSCNN-D model for hyperspectral image reconstruction

The hyperspectral hypercube and RGB images were fed into an advanced CNN-based hyperspectral image reconstruction method referred to as HSCNN-D (Shi et al., 2018). This method employs a densely connected structure (Fig. 1) to reconstruct hyperspectral data from RGB images. HSCNN-D replaces the residual block of another reconstruction algorithm called HSCNN-R to provide a more precise solution by increasing the network depth. A path-widening fusion constructed around the dense structure is also incorporated to improve the model's performance. The dense structure can significantly mitigate gradient vanishing during training. The dense connection allows the k-th layer to get as input the features from all antecedent layers (i.e., $f_0, f_1, ..., f_{k-1}$), as shown in Eq. 2. Thus, the propagation issue caused by the network's growing profundity can be resolved. Along with the intrinsic benefit of convergence, the compact structure possesses a second quality that makes it appropriate for hyperspectral reconstruction. Given an RGB image as input, it is possible to reconstruct a hyperspectral image with multiple channels as output. The key distinction between input and output is a reduction in channels. The concatenation operator implemented in each dense block increases the channel count, which can produce a more accurate model for this reconstruction problem.

$$f_k = c_k([f_0, f_1, ..., f_{k-1}]) \tag{2}$$

Where, $c_k(.)$ refers to the k-th convolution layer and $[f_0, f_1, ..., f_{k-1}]$ denotes the concatenation of the features generated by preceding layers.

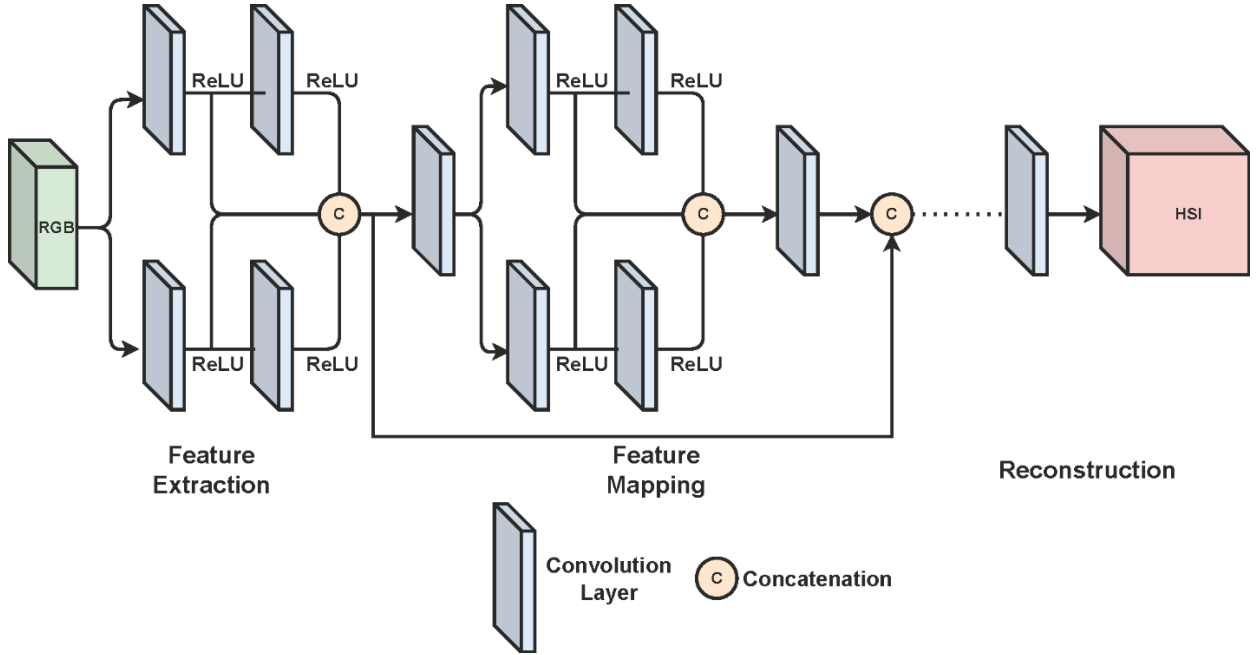

Fig. 1: Structure of the HSCNN-D method.

2.6.4 Reconstructed image evaluation

The image reconstruction model's efficacy was evaluated using the three-evaluation metrics MRAE, RMSE, and PSNR (Wei et al., 2019). Eq. 3-5 symbolize the computations involved in carrying out the metrics. The model was trained until no discernible decline in training MRAE loss. The performance of the training model was also assessed for the validation and test set.

$$MRAE = \frac{1}{n}\sum_{i=1}^{n}\left(\left|I_{rc}^{(i)} - I_{gt}^{(i)}\right|/I_{gt}^{(i)}\right) \tag{3}$$

$$RMSE = \sqrt{\frac{1}{n}\sum_{i=1}^{n}\left(I_{rc}^{(i)} - I_{gt}^{(i)}\right)^2} \quad (4)$$

$$PSNR = 10 \times \log_{10}\left(\frac{255^2 \times n}{\left\|I_{rc} - I_{gt}\right\|^2}\right) \quad (5)$$

Where $I_{gt}$ and $I_{rc}$ represent ground truth and reconstructed hyperspectral images, respectively, and n denotes the total pixels.

2.6.5 Spectra extraction from reconstructed images

The trained model was used to reconstruct all 282 images fed to the network. Each reconstructed image had the same dimension as the ground truth images (512×512× λ). The ROI generated in section 2.4 was again imposed on the reconstructed images. The reflectance spectra from the ROIs were calculated by averaging all pixel spectra within the ROI to generate a single mean spectrum for each image. As two images were attributed to one sample, the spectra for both images were averaged. Thus, 141 spectra were acquired for ground truth and reconstructed hyperspectral images. These spectra were finally analyzed for quality predictions of the samples.

2.7 Spectral analysis

2.7.1 Spectral preprocessing

In spectroscopic sensing, factors like light scattering, instrumental drift, baseline shift, and slope variation can affect spectral data, necessitating pre-processing before calibration (Cozzolino et al., 2023). Data pre-processing aims to eliminate or minimize variability or physical effects unrelated to the property of interest. Effective spectroscopic analysis depends on proper pre-processing. While correct pre-processing can enhance modelling accuracy, incorrect choices can lead to inaccurate predictions. Selecting the right method is challenging due to varying dimensionality and noise conditions in spectral data sets. In this study, three pre-processing methods were used: multiplicative scatter correction (MSC), standard normal variate (SNV), and Savitzky–Golay smoothing with the first derivative (SG-1st der). The choice of techniques was based on the performance of the full spectrum model.

2.7.2 Spectral model development

The partial least squares regression (PLSR) model was used to create the calibration model in the full spectral range on the selected features and the spectra extracted from the reconstructed images. PLSR is regarded as the most prevalent multivariate chemometric technique (Ragni et al., 2012). It effectively manages highly collinear and chaotic data and identifies spectral regions with the greatest information for predicting reference values. In PLSR, the spectral matrix (X) and the reference variable (Y) are mapped onto the subspace of latent variables (LVs) to calculate the maximal covariance (Wold et al., 2001). In contrast to the highly interconnected original variables, these latent variables are informative and mutually exclusive. A minimum number of LVs utilized in the PLSR model to minimize complexity and generate a practical model resistant to underfitting and overfitting (Wang et al., 2022). Therefore, the optimal number of LVs is required to develop accurate and reliable prediction models. Numerous researchers have utilized the least value of the root mean square error of cross-validation (RMSECV) to select the best LVs (Wang et al., 2022). The number of latent variables was identified using the least value of RMSECV using the leave-one-out cross-validation (LOOCV) technique (Zhu et al., 2017). The developed model was used to predict all the quality attributes of the samples.

2.7.3 Spectral model evaluation

Several statistical parameters, including coefficients of determination for calibration ($R_c^2$), root mean square error of calibration (RMSEC), coefficients of determination for validation ($R_v^2$), root mean square error of validation (RMSEV), coefficients of determination for prediction ($R_p^2$), root mean square error of prediction (RMSEP), and the ratio of prediction to deviation (RPD), were determined to evaluate the accuracy of the calibration models. A robust model possesses high correlation coefficients ($R_c^2$, $R_v^2$, and $R_p^2$), a high RPD, and low errors (RMSEC, RMSEV, and RMSEP) (Wang et al., 2022).

2.8 Computational environment

The computations in this investigation were carried out using Google Colaboratory Pro (Colab Pro), a cloud computing platform (Google LLC, Mountain View, CA, USA). The Colab Pro virtual machine featured a two-core Intel(R) Xeon(R) CPU @ 2.30GHz and 25 GB of RAM and an extended lifetime, and a Tesla P100 GPU (NVIDIA, Santa Clara, CA, USA) with 16 GB RAM. The analyses were performed in Python 3.9. Open-source Python libraries (McCann et al., 2022), such as Scikit-learn, PyTorch, and OpenCV, were used to train the reconstruction model. It took about 34 hours to train the image reconstruction model.

3. Results and discussion

3.1 Calibration model on the full spectra

After the spectra acquisition from the ground truth hyperspectral images, the PLSR calibration model was developed using the whole spectral range. The spectral dataset was randomly partitioned into a calibration (60%), validation (20%), and prediction (20%) set. In spectroscopy, random sampling is a commonly used method to select a sample from a population (Shenk and Westerhaus, 1991). A total of 85 samples were used to develop the calibration, while 28 samples each were used to validate and predict the effectiveness of the model. The response variable, SSC, had a mean of 11.74 ± 1.54 ºBrix for the training set, a mean of 11.56 ± 1.85 ºBrix for the validation set and a mean of 11.74 ± 1.57 ºBrix for the prediction set. The number of LVs was chosen based on the minimal RMSECV value by applying the leave-one-out cross-validation (LOOCV) technique (Zhu et al., 2017).

Table 1 displays the efficacy of the PLSR model across the entire spectral spectrum. In addition to the raw spectra, various spectral pre-processing techniques, such as MSC, SNV, and SG-1st derivative, were utilized. However, the PLSR model developed from the raw spectra outperformed PLSR models developed from other pre-processed spectra, as shown in Table 1.

Table 1: PLSR model with different pre-processing on the whole spectral range (400-1000 nm)

| Pre-processing | LV | Calibration | | Validation | | Prediction | | |
|---|---|---|---|---|---|---|---|---|
| | | $R_c^2$ | RMSEC (ºBrix) | $R_v^2$ | RMSEV (ºBrix) | $R_p^2$ | RMSEP (ºBrix) | RPD |
| RAW | 7 | 0.6873 | 0.8542 | 0.5851 | 0.9945 | 0.5168 | 1.2596 | 1.4385 |
| MSC | 6 | 0.6933 | 0.846 | 0.5814 | 0.9989 | 0.45 | 1.3438 | 1.3484 |
| SNV | 6 | 0.6929 | 0.8465 | 0.5806 | 0.9998 | 0.4546 | 1.3381 | 1.3541 |
| SG-1st derivative | 6 | 0.7108 | 0.8214 | 0.4795 | 1.1139 | 0.3965 | 1.4076 | 1.2872 |

Some studies have also applied chemometric models to predict the SSC content of sweet potatoes using spectroscopy. Reyes et al. (2018) used spectroscopic analysis to achieve $R^2$ of 0.3616. Shao et al. (2020) applied HSI and PLSR to predict the SSC of 'Red Banana' sweet potatoes. They achieved $R_c^2$, RMSEC, $R_p^2$, RMSEP, and RPD of 0.77, 0.31, 0.73, 0.31 and 1.88, respectively, on the raw spectra with 14 LVs.

3.2 Feature selection from the ground-truth spectra

The primary limitation of HSI lies in the system's speed, which is influenced by the size of the images. The processing, display, and analysis of hyperspectral data require significant time, impeding real-time applications. Nevertheless, HSI can serve as a valuable tool in developing a multispectral imaging system by selecting crucial bands while maintaining the overall accuracy of the hyperspectral system intact (Wang et al., 2022). Conversely, not all spectral bands within hyperspectral images contain significant information. Certain bands may contain redundant, irrelevant, or useless data (Pu et al., 2015). Therefore, an effective approach involves selecting pertinent information while discarding redundant, irrelevant, and useless data.

GA is a computational technique inspired by the mechanisms of natural evolution that can effectively optimize wavelength combinations for spectroscopic analysis (Costa et al., 2020). In the present study, GA was employed to select a total of 15 important wavelengths (403, 476, 525, 528, 608, 610, 661, 700, 715, 733, 811, 838, 930, 942, and 991 nm). Some studies selected important wavelengths to determine the SSC of sweet potatoes. Shao et al. (2020) selected 35 wavelengths using SPA method ranging from 400 to 1000 nm to predict the SSC of 'Red banana' sweet potatoes. Sanchez et al. (2020) applied two laser wavelengths (658 and 780 nm) to determine the SSC content of sweet potatoes.

3.3 Training and validation of image reconstruction model

The selected important wavebands were extracted from each hyperspectral image to prepare the data for developing the image reconstruction model. These hyperspectral images with the chosen wavebands (ground-truth images) worked as the label information, whereas the respective RGB images worked as the input to the model. The Adam solver for optimization (He et al., 2015) was used to train the model, with the momentum factor adjusted to 0.90 and the coefficient of weight decay (L2 -Norm) set to 0.0001. The batch size was 8, the patch size was 128, the stride was 8, and the initial learning rate utilized to train the model was $2e^{-4}$. In each epoch, the learning rate decreased exponentially. The algorithm proposed by He et al. (2015) is employed to initialize the weights except for the final layer. In particular, the final reconstruction layer is initialized with random weights derived from Gaussian distributions with σ=0.001. Each convolutional layer's biases are initialized to zero.

After initializing all the hyperparameters, the model was trained for 50 epochs. Each epoch contained 1000 iterations. The MRAE loss function was used as the key criteria for the computation of the model. MRAE is more stable against outliers and is more uniform in its treatment of wavebands under various lighting conditions (Shi et al., 2018). Additionally, models trained with the MRAE loss function converge faster and in fewer epochs (Zhao et al., 2020). Fig. 2 shows the training MRAE change based on the epoch number. The lowest training MRAE (0.7700) was recorded on the 36$^{th}$ epoch. The MRAE showed an incline for the following epochs. Thus, the model with the 36$^{th}$ epoch was considered for evaluating the validation and test set. The evaluation of the validation and test set is listed in Table 2. The lower MRAE, RMSE, and higher PSNR suggest that the model was optimized enough to reconstruct the images properly.

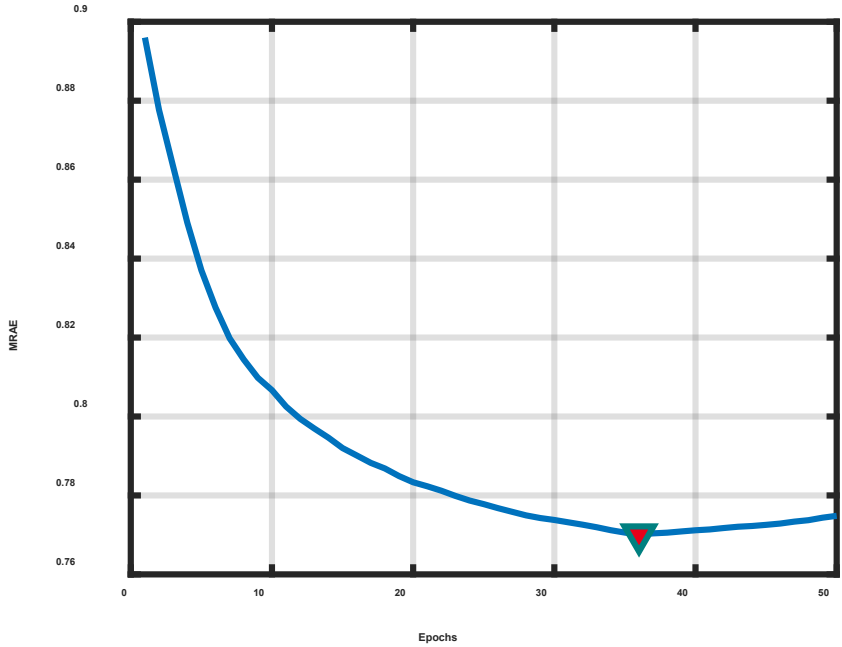

Fig. 2: Training MRAE loss of image reconstruction model.

Table 2: Evaluation metrics of the image reconstruction model on validation and prediction set

| Data | MRAE | RMSE | PSNR (dB) |
|---|---|---|---|
| Validation set | 0.8601 | 0.0545 | 26.8648 |
| Prediction set | 0.7914 | 0.0566 | 25.905 |

3.4 Visual and spectral quality assessment

Following the successful execution of the model, a comparative analysis was conducted to assess the visual and spectral quality of the reconstructed hyperspectral images in relation to the ground-truth images. Fig. 3 illustrates the ground-truth and reconstructed images of selected spectral wavelengths. It is evident from Fig. 3 that the reconstructed images closely resemble the ground-truth images when examined visually. However, slight deviations in the pixel intensities of the reconstructed images are noticed, especially in the short-wave near-infrared (SW-NIR) region (700-1000 nm), as indicated in the figure. Furthermore, the spectral quality of the reconstructed images was evaluated. Fig. 4 (a-c) presents the ground-truth and reconstructed spectra for three samples, each possessing distinct reference values. It can be observed from Fig. 4 that the reconstructed spectral magnitude values and trends closely align with those of the ground-truth spectra. The spectral characteristics exhibit an exact match within the visible spectral range. However, slight deviations are -noticeable in certain wavebands within the SW-NIR region for specific samples. Nevertheless, the overall spectral pattern of the reconstructed images consistently follows that of the ground-truth images.

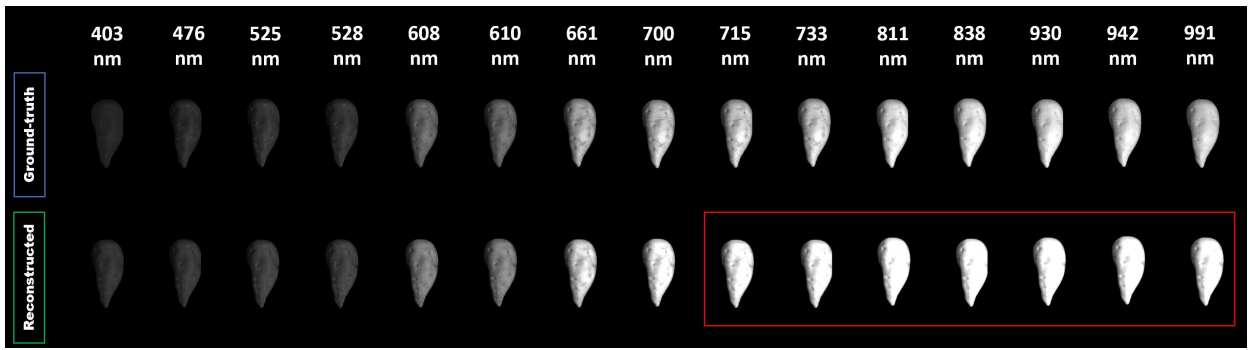

Fig. 3: Ground-truth and reconstructed images of the sweet potato samples. As indicated, some minor pixel inconsistencies are noticed in the reconstructed images compared to the ground-truth images in the SW-NIR (700-1000 nm) regions.

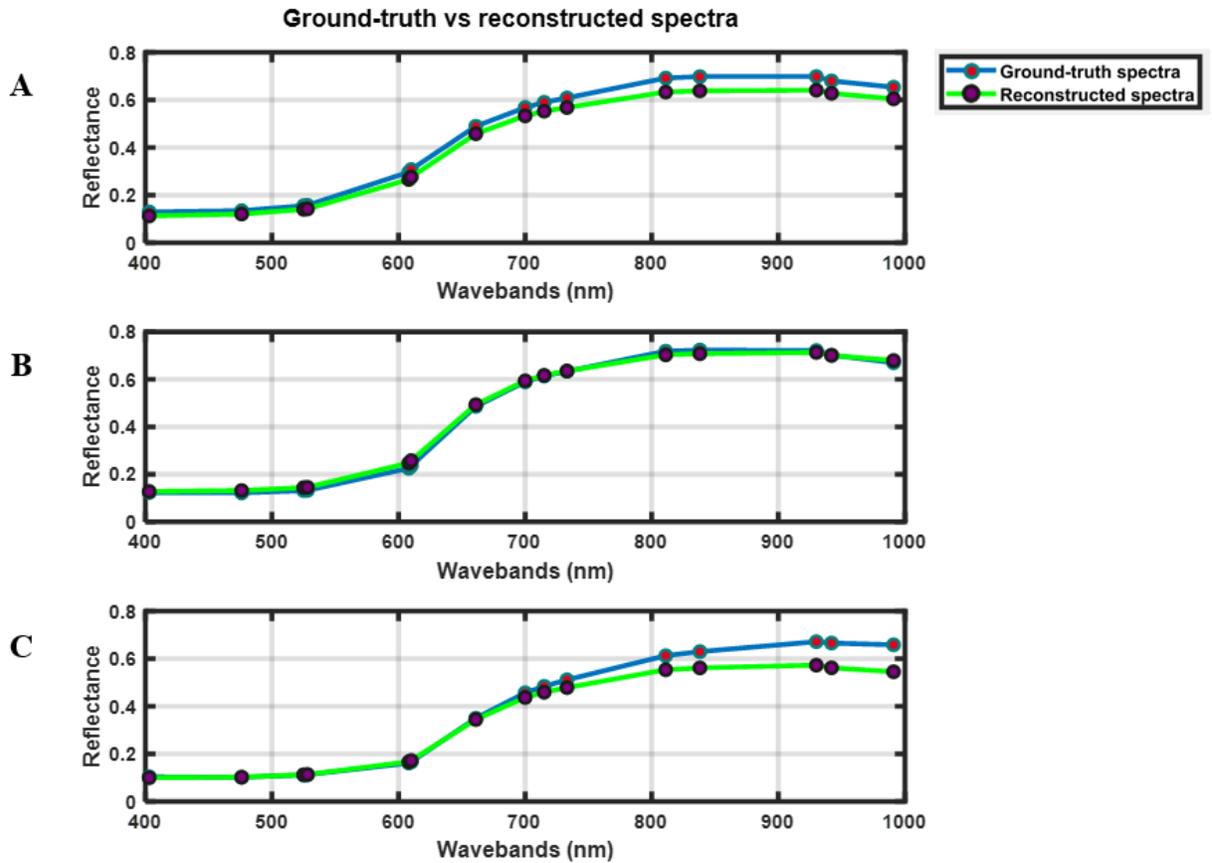

Fig. 4: Assessment of the reconstructed spectra (a) Reconstructed spectra for the sample with minimum (9.3 °Brix) reference value, (b) Reconstructed spectra for the sample with average (11.7 °Brix) reference value, (c) Reconstructed spectra for the sample with maximum (17.5 °Brix) reference value.

3.5 Calibration models using selected wavelengths from ground truth and reconstructed spectra

Following the spectra extraction from the reconstructed hyperspectral images, PLSR models were again developed using the ground-truth and reconstructed spectra. The effectiveness of these PLSR models in predicting the SSC content is displayed in Table 3. Even though 92.65% of the feature wavebands were removed, the later models outperformed the models with full wavebands. These models demonstrated improved model quality by having higher $R^2$ and RPD, lower RMSE values, and a smaller difference between the RMSE values compared to the model applied on the full wavebands. These outcomes are not unexpected. Most co-linearity issues between variables were resolved once the unnecessary variables were removed from the calibration information set, leading to a stronger model regarding accuracy and resilience. So, the selected wavebands were important in predicting the SSC content of the samples.

The PLSR model developed on the reconstructed spectra showed a great performance and outperformed the PLSR model on the full spectral range with an $R_c^2$ of 0.5516, an RMSEC of 1.0903 ºBrix, an $R_v^2$ of 0.7817, an RMSEV of 0.6593 ºBrix, an $R_p^2$ of 0.6932, RMSEP of 0.9109 ºBrix, and an RPD of 1.8054. However, the PLSR model on the ground-truth spectra with selected feature wavebands slightly outperformed the PLSR model on the reconstructed spectra with an RPD of 1.8221. The LOOCV technique was once again used to find the lowest RMSECV values to optimize the LVs for both models. Some studies identified important wavelengths and applied chemometric techniques to predict the SSC content of sweet potatoes. Sanchez et al. (2020) used a laser backscattering imaging technique to achieve $R_c^2$, RMSEC, $R_p^2$, and RMSEP of 0.26, 0.32, 0.18, and 0.33, respectively, for the wavelength of 658 nm. Shao et al. (2020) applied SPA-PLSR technique on the 35 selected wavelengths to register $R_c^2$, RMSEC, $R_p^2$, RMSEP and RPD of 0.74, 0.34, 0.75, 0.38, and 1.51, respectively. No studies have reported the prediction of SSC of sweet potatoes based on their reconstructed spectra to date. However, Zhao et al. (2020) applied HSCNN-R algorithm to reconstruct the hyperspectral image and assess the SSC of tomatoes. They obtained an $R^2$ of 0.51 for predicting the SSC of tomatoes based on reconstructed spectra.

Table 3: PLSR model on the ground-truth and reconstructed spectra for the selected feature wavelengths for the prediction of SSC.

|  | | Calibration | | Validation | | Prediction | | |
| --- | --- | --- | --- | --- | --- | --- | --- | --- |
| Spectra | LV | $R_c^2$ | RMSEC (°Brix) | $R_v^2$ | RMSEV (°Brix) | $R_p^2$ | RMSEP (°Brix) | RPD |
| Ground-truth | 4 | 0.5572 | 1.0834 | 0.6437 | 0.8423 | 0.6988 | 0.9025 | 1.8221 |
| Reconstructed | 4 | 0.5516 | 1.0903 | 0.7817 | 0.6593 | 0.6932 | 0.9109 | 1.8054 |

4. Conclusions

This study presents the practicality of hyperspectral image reconstruction as a valuable approach for assessing agro-product quality, using sweet potatoes as the focal point. Initially, spectra were extracted from segmented hyperspectral images acquired within the visible near-infrared range. Subsequently, applying the genetic algorithm aided in identifying crucial wavelengths essential for accurate assessments. Notably, the study successfully implemented the HSCNN-D hyperspectral reconstruction algorithm for the first time for agro-product quality assessment, utilizing the selected wavebands as training labels and RGB images as inputs. The trained model showed impressive accuracy in reconstructing hyperspectral images and spectra, exhibiting similarity with the ground-truth spectra. Furthermore, the PLSR model applied to the reconstructed spectra outperformed the model applied to the full spectral spectra collected from HSI, highlighting its potential for precise quality evaluation. This research opens up promising possibilities, indicating the feasibility of generating hyperspectral images of sweet potatoes and other agricultural products using consumer-level cameras while encouraging exploration into other valuable hyperspectral properties of interest.